\documentclass{article}
\usepackage{ijcai23}

\usepackage{times}
\usepackage{soul}
\usepackage{url}
\usepackage[hidelinks]{hyperref}
\usepackage[]{inputenc}
\usepackage[small]{caption}
\usepackage{graphicx}
\usepackage{amsmath}
\usepackage{amsthm}
\usepackage{booktabs}
\usepackage{algorithm}
\usepackage{algpseudocode}
\usepackage[switch]{lineno}
\usepackage{color}
\usepackage{booktabs}
\usepackage{subfigure}
\usepackage{multirow}
\usepackage{makecell}

\usepackage{titlesec}
\newcommand{\fig}[3]{
    \begin{figure}[tb]\centering
    \includegraphics[width=\columnwidth]{fig/#1}
    \caption{#2}
    \label{#3}\end{figure}
    {}}

\algnewcommand{\LineComment}[1]{\Statex \(//\) #1}

\title{TTSWING: a Dataset for Table Tennis Swing Analysis}

\author{
Che-Yu Chou$^{1\dagger}$\and
Zheng-Hao Chen$^{1\dagger}$\and
Yung-Hoh Sheu$^{2}$\and
Hung-Hsuan Chen$^{1}$\And
Sheng K.~Wu$^{3}$
\affiliations
$^1$Department of Computer Science and Information Engineering, National Central University\\
$^2$Department of Computer Science and Information Engineering, National Formosa University\\
$^3$Department of Sport Performance, National Taiwan University of Sport\\
\emails
\ 
tetsuyu89617@gmail.com,
s10727220@cycu.org.tw,
yhsheu@nfu.edu.tw,
hhchen1105@acm.org,
skwu@ntus.edu.tw 
}

\begin{document}

\maketitle
\def\thefootnote{$\dagger$}\footnotetext{The first two authors contributed equally to this paper.}

\begin{abstract}

We introduce TTSWING, a novel dataset designed for table tennis swing analysis. This dataset comprises comprehensive swing information obtained through 9-axis sensors integrated into custom-made racket grips, accompanied by anonymized demographic data of the players. We detail the data collection and annotation procedures. Furthermore, we conduct pilot studies utilizing diverse machine learning models for swing analysis. TTSWING holds tremendous potential to facilitate innovative research in table tennis analysis and is a valuable resource for the scientific community. We release the dataset and experimental codes at \url{https://github.com/DEPhantom/TTSWING}.

\end{abstract}

\section{Introduction}

Since its inclusion as an Olympic sport in 1988, table tennis has gained widespread popularity and is enjoyed worldwide not only as a competitive sport but also as a common recreational pastime among players of all levels and ages.  Meanwhile, with the advances in sensor technologies and machine learning algorithms, there is an increasing interest in using data-driven approaches to analyze sports, including table tennis. Such approaches provide valuable insights into player performance and inform training programs for players. Additionally, these approaches help to identify biomechanical factors that may increase the risk of injury and inform injury prevention strategies.

This paper introduces TTSWING (Table Tennis Swing Dataset), a novel dataset that includes swing information gathered by the 9-axis sensors located in the grips of customized paddles. The swing here refers to the powerful forehand smash movement.  In addition to the swing information, the dataset also includes players' anonymized demographic information, such as age, gender, height, weight, and years of playing experience. This comprehensive dataset provides a valuable resource for studying table tennis swings and can be used to develop new techniques and approaches for improving table tennis skills. 

Based on the collected dataset, we perform pilot studies on swing analyses using various machine learning models. Specifically, we investigate the feasibility of using machine learning techniques to automatically predict a player's age, gender, playing experience in years, racket-holding hand, and swing mode, which includes swinging in the air, full power stroke, and stable hitting. Our results demonstrate that the dataset is suitable for training and evaluating machine learning models for table tennis swing analysis. We also provide a performance comparison for different machine learning models, which can be a starting point for future research. We believe that our dataset has the potential to enable new studies in the field of table tennis analysis and will be a valuable resource for the community.

In summary, our work has the following contributions.

\begin{itemize}
    \item We present a new dataset, TTSWING, that captures professional table tennis players' swings along with anonymized demographic information of players. We describe the collecting and annotating process and conduct initial studies on the dataset.  To the best of our knowledge, this is the largest open dataset for professional players' swing information with their anonymized demographic information.

    \item We show that our dataset can be used to train and evaluate machine learning models for swing analysis, opening up new avenues for research in this area. 

    \item We release the experimental codes at \url{https://github.com/DEPhantom/TTSWING} for reproducibility. These codes and the open dataset can serve as a foundation for future research. As a result, our study will facilitate research in table tennis analysis.
\end{itemize}

The paper is structured as follows. In Section~\ref{sec:rel-work}, we review previous studies on using sensors to collect information from table tennis players. Section~\ref{sec:data} outlines the methodology we employed to collect the data, including the types of sensors used, the number of players involved, and other relevant details. Section~\ref{sec:analysis} presents the analysis of the collected dataset through pilot studies. Finally, in Section~\ref{sec:disc}, we discuss the potential implications of our research and suggest future research directions.
\section{Related Work} \label{sec:rel-work}

Advancements in sensor technology have allowed for the collection of detailed information on the performance of table tennis players. Consequently, it is possible to utilize sensors to measure various aspects of a table tennis player's performance. This section reviews previous studies on using sensors to collect table tennis performance.

Several research papers have employed wearable sensors directly attached to the players' bodies to measure their motion data.  For example, in ~\cite{ren2019kinematic}, the authors attached a DELSYS sensor to ten points on the right arm to collect information on muscle, including speed, joint angle, and root mean squares of acceleration. The authors analyzed the differences between professional and amateur players when stroking.  A follow-up paper expanded the range of data collection by placing a hang3.0 sensor on nine different areas, e.g., hands, limbs, and waists~\cite{yanan2021using}. The authors classified and clustered similar motions based on acceleration and angular velocity and used the results to improve the stroke posture of players. While multiple sensors may collect more information, using excessive portable devices may make players' movements less natural.

Modern smartphones and smart watches commonly contain internal sensors.  As such, they are convenient tools for motion detection.  Some studies have placed the smart device on the wrist of the player to collect acceleration and angular velocity data during the stroke~\cite{viyanon2016swingpong,sharma2018wearable,ferreira2022classification}. These data are then used to predict stroke type and analyze shot spots. However, this approach requires the device to be worn on the hand holding the racket, which may indirectly impact the player's performance. This also means we cannot obtain data close to what the player produces during an actual game.

Some studies have embedded multiple shock sensors inside a table tennis racket~\cite{blank2016ball}. This approach predicts the shot spot on the racket based on the interaction of multiple sensors.  However, this method still makes errors in positioning, especially at partial racket edges.  As a result, it is difficult to complete the dataset and analyze the recorded data.

Another method for detecting stroke motion is to embed sensors into the racket grip~\cite{boyer2013low}.  Some follow-up works further used this method to predict the spin of table tennis and stroke type~\cite{blank2015sensor,blank2017ball}. Compared to other collection methods, embedding sensors into the grip reduces the burden on the player and makes data collection easier. Additionally, the sensor embedded in the grip captures the motion of the racket, allowing for the detection of even subtle changes in racket movement. 

We have chosen to utilize the last method for data collection in our paper. However, previous studies only collected the swing information from a limited number of players, typically ranging from a few to a dozen~\cite{boyer2013low,blank2015sensor,blank2017ball}.  In contrast, our dataset is a unique resource for future research as we collected and analyzed information from nearly 100 elite players, providing a much larger sample size than past experiments.
\section{The TTSWING Dataset} \label{sec:data}

\subsection{Challenges of Collecting Swing Data}

Collecting data on table tennis swings presents a range of challenges. Commonly used methods, such as video recording or attaching sensors to the human body, have limitations. Video recording requires a fixed camera setup.  However, it can be challenging to replicate the same environmental setup from one place to another, making it challenging to collect accurate data in different environments.  On the other hand, attaching sensors to the body using smartwatches, smartphones, or other sensors may influence the players' movements, which can create interfering factors in the analysis. Embedding sensors in the equipment, such as the paddle, is a better option. However, previous studies using this approach only collected data from a small number of players, typically no more than 20. Moreover, it requires significant manpower to split continuous signals collected by sensors into stroke-based data.

In light of these limitations, we develop a new method that addresses the challenges of collecting table tennis swing data. We embed sensors directly into the racket to collect data from over 90 professional players, generating a dataset of more than 90,000 strokes. Our approach allows for the accurate collection of stroke information without significantly affecting players' performance. Additionally, we have developed an automated method to split the collected continuous waveform data into each stroking data, making data processing more efficient. By collecting data from a large number of players, our method provides a more comprehensive understanding of the mechanics and nuances of table tennis swings. Our approach overcomes the limitations of previous studies, which only collected data from a small number of players. The resulting dataset opens up new possibilities for research and development in the field, allowing for the creation of more advanced applications and complex models.

\subsection{Hardware} \label{sec:hardware}

\fig{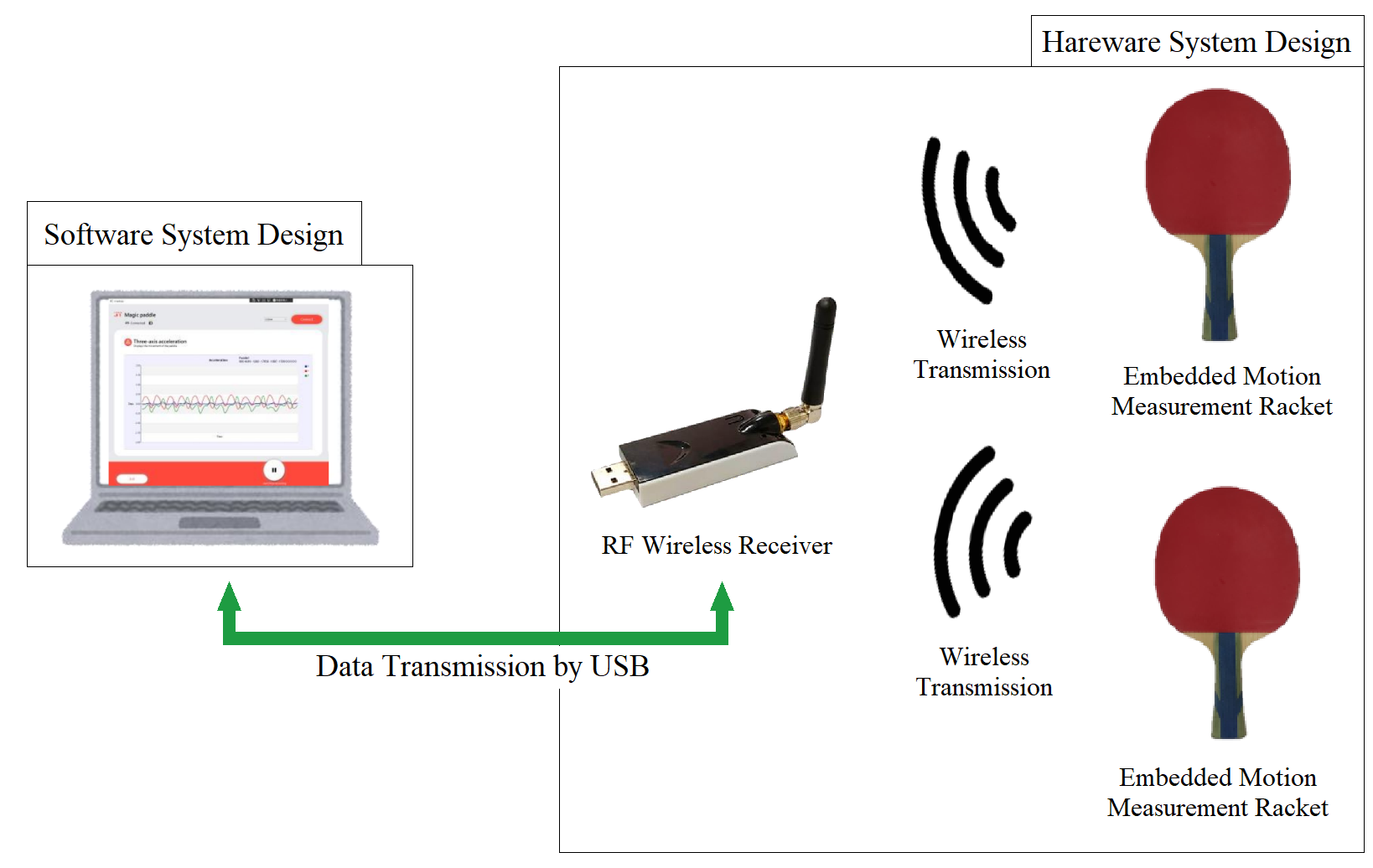}{An overview of the entire system.}{fig:entire-design}

\fig{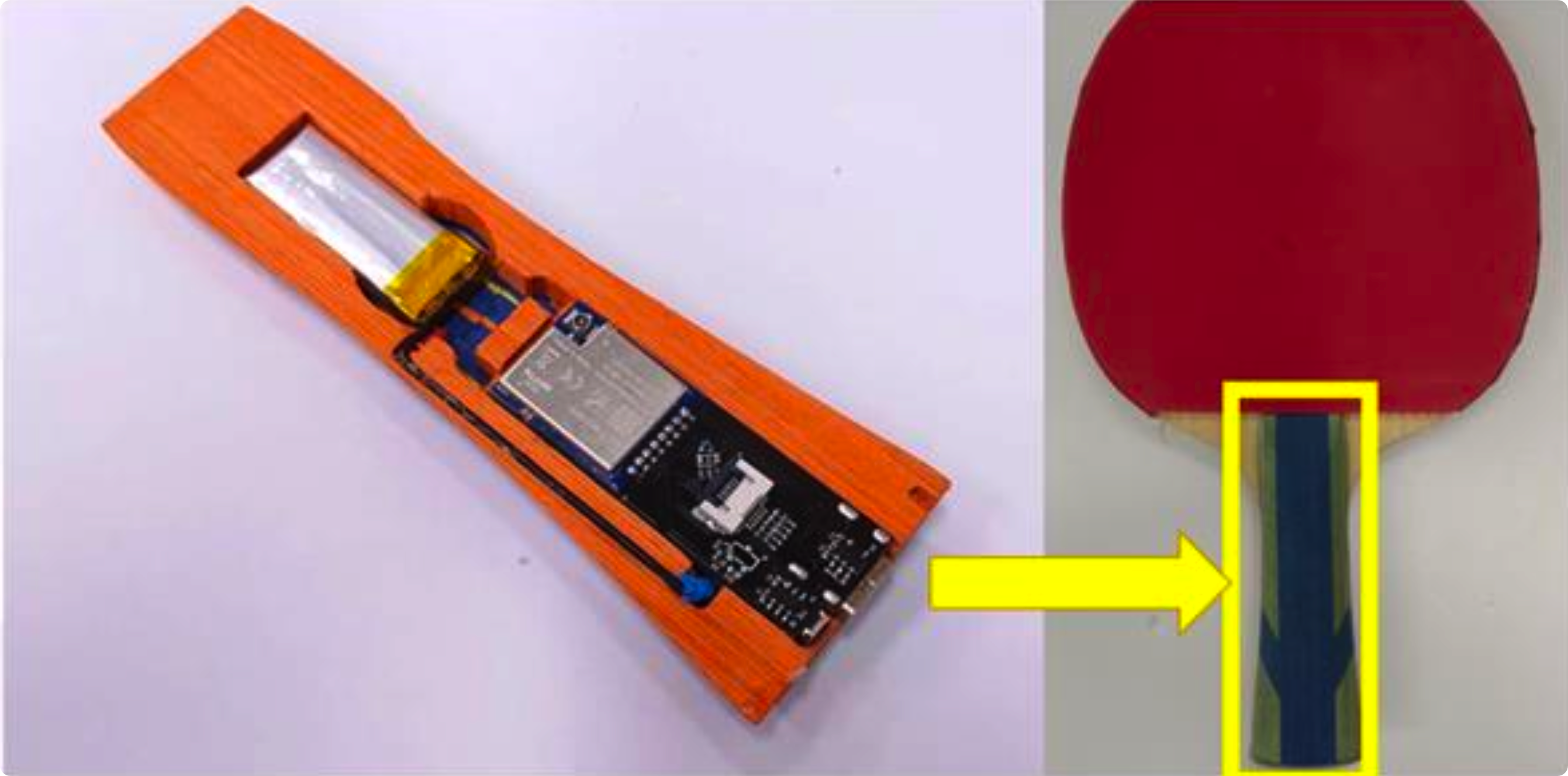}{Embedding detection sensors and communication modules inside the paddle of a racket.}{fig:circuit-board}

\fig{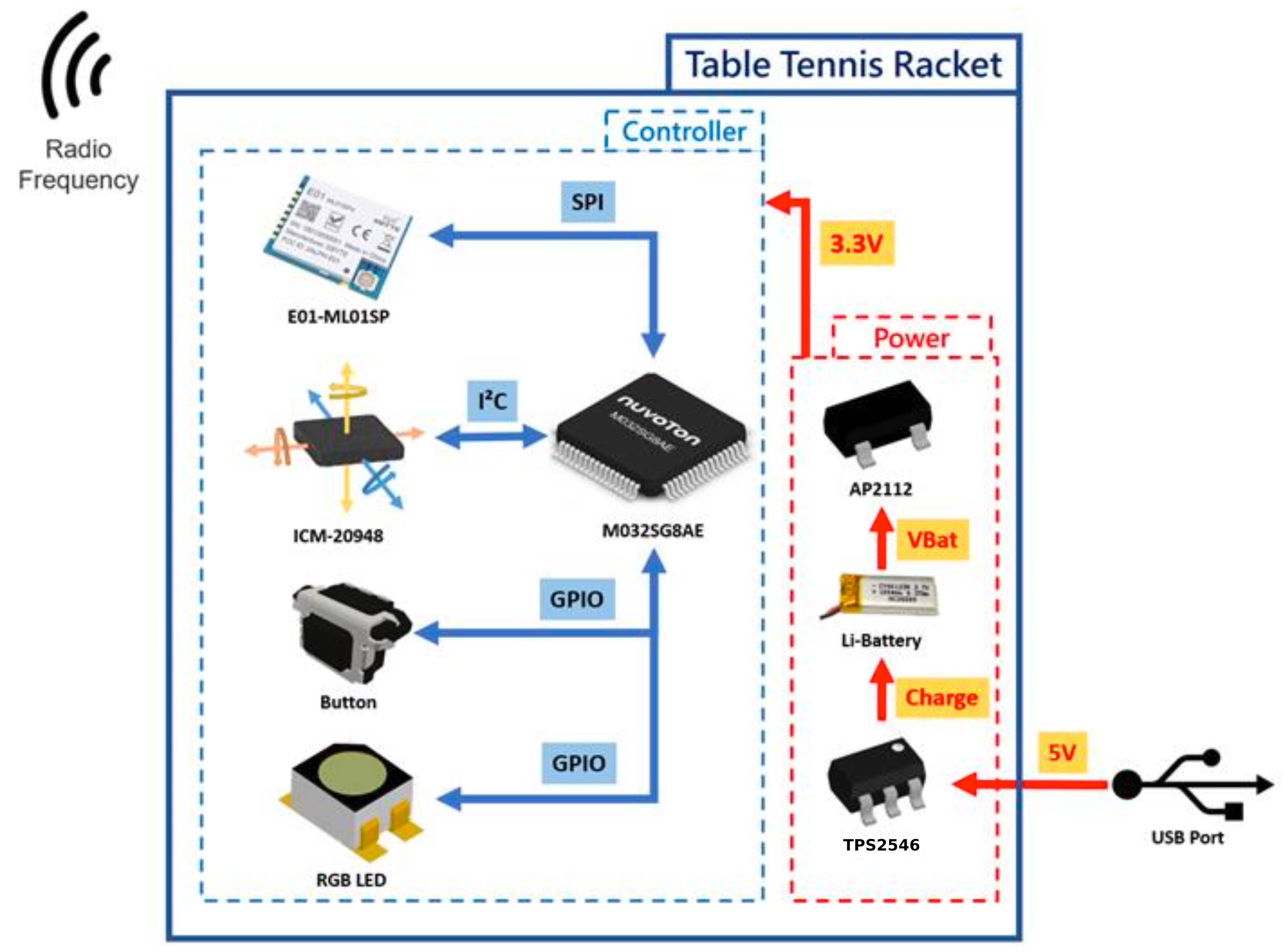}{Hardware architecture of the sensor in the paddle of a racket.}{fig:racket-arch}

\fig{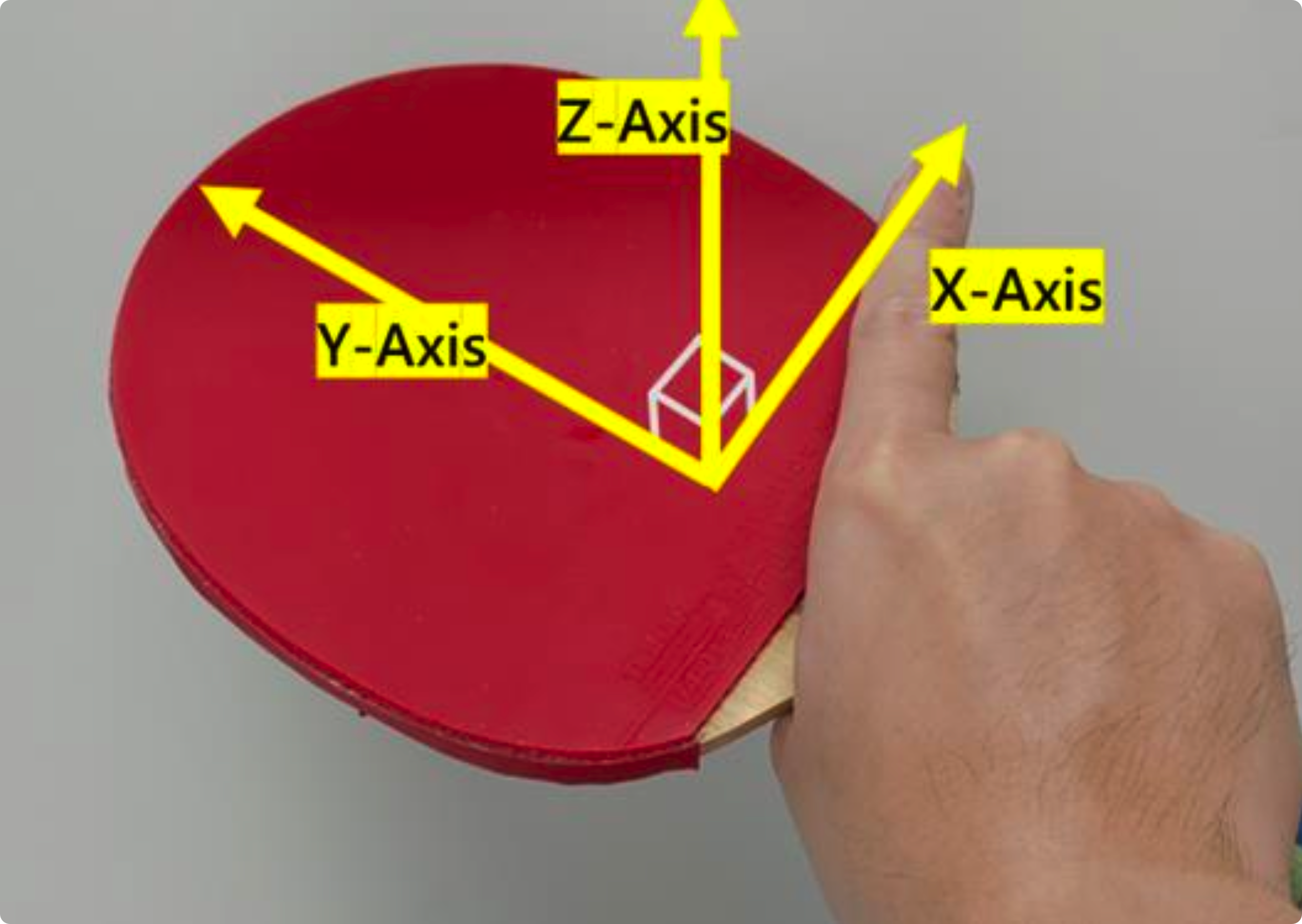}{Definitions of the three axes.}{fig:direction-def}

Figure~\ref{fig:entire-design} gives an overview of the entire system. In order to collect data, motion sensors are embedded into the racket grips.  The collected data are transferred to the RF wireless receiver, which further transfers the information to a computer via a USB port.

Figure~\ref{fig:circuit-board} shows how we embed
the hardware in the table tennis racket.  The embedded components include an inertial measurement unit (ICM-20948), a module for radio frequency (RF) wireless transmission (E01-ML01SP), and affiliated components such as the button and the RGB LED for simple I/O communications, as shown in Figure~\ref{fig:racket-arch}.  The system is powered by a lithium battery, and a TPS2546 USB charging port is connected to a 5-V DC external power supply to maintain the battery's power.  Eventually, the embedded racket weighs approximately 190 grams, falling within the weight range of a regular racket.

The motion sensor, ICM-20948, is the key component in collecting the swing data.  The sensor integrates a 3-axis accelerometer, 3-axis gyroscope, and 3-axis compass, forming a 9-axis sensor that can effectively measure 3-axis acceleration, 3-axis angular velocity, and 3-axis magnetic field data. The three axes are defined according to Figure~\ref{fig:direction-def}: the positive x-axis is to the right, the positive y-axis is forward, and the positive z-axis is perpendicular to the red side of the racket.

\subsection{Swing Data Collection} \label{sec:data-collection}

We invited 93 Taiwanese players from Group A to participate in the data collection process.\footnote{Players in Group A are either elite players who have majored in physical education or have won medals in important competitions.} The participants were asked to perform at least one of three different modes of swings using the proposed racket.  The three modes include swing in the air, full power stroke, and stable hitting. Each mode requires the participants to swing the racket 50 times continuously to generate a complete waveform set. In the full power stroke mode, there are three different ball speeds set by the serving machine for the players to hit.

\subsection{Split the Complete Waveform Set into Separated Stroke Waveforms}

\begin{algorithm}[tb]
    \caption{Waveform split algorithm}
    \label{alg:waveform-split}
    \textbf{Input}: 3-axis acceleration, angular velocity, magnetic field
    \textbf{Output}: Waveform for each stroke
    \begin{algorithmic}[1] 
        \State Compute the integrated waveform by Equation~\ref{eq:combine-singal};
        \LineComment Normalize integrated waveform
        \State Calculate the trend line by the combined signal;
        \State Subtract the trend line from the combined signal;
        \State Cutoff high frequency by a low-pass filter;
        \State Scale the signals to be within the range $[0, 1]$;
        \LineComment Segment the waveform
        \While{current number of intersection points$\times$2 $<$ total number of stroke} 
        \State Move down the horizontal line;
        \EndWhile
        \State Search for troughs to the right and left;
        \State Split original waveform by troughs;
        \State \textbf{return} The waveform for each stroke;
    \end{algorithmic}
\end{algorithm}

\fig{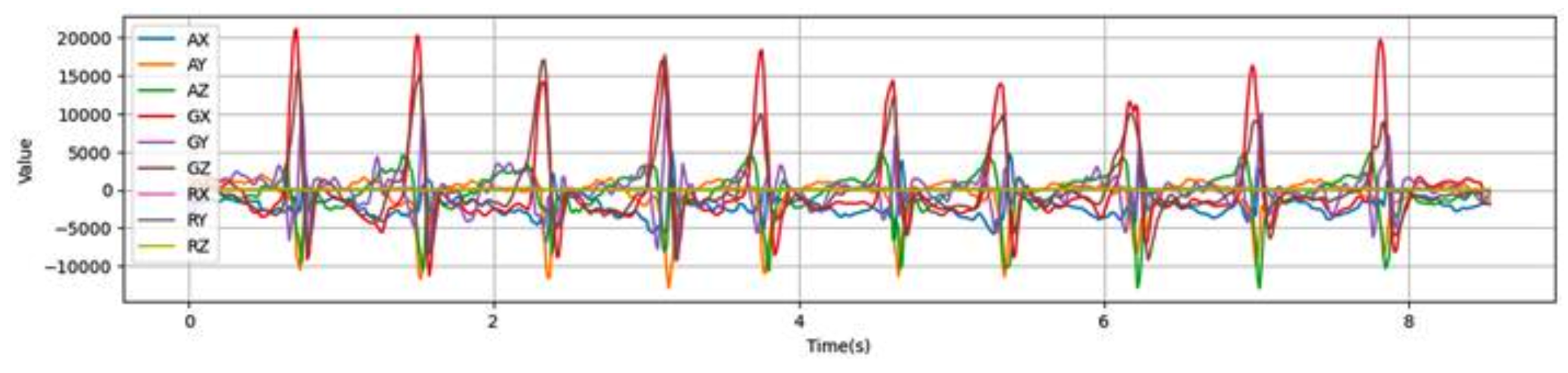}{Raw data of continuous stroke.}{fig:waveform-data}

This section details the methodology employed to divide a raw waveform set into stroke-based waveforms.  Algorithm~\ref{alg:waveform-split} shows the pseudocode of the splitting process.  We describe the details below.

As mentioned in Section~\ref{sec:data-collection}, each participant in the study swung the racket 50 times continuously, resulting in a complete waveform set. We would like to divide each complete waveform set into 50 separate stroke waveforms for further analysis. However, this proved to be a challenging task since different strokes exhibit distinct strengths and trajectories, leading to the generation of unique waveforms. Figure~\ref{fig:waveform-data} illustrates a portion of a complete waveform set comprising 10 consecutive strokes, with each stroke waveform displaying a similar but distinct shape. 

\fig{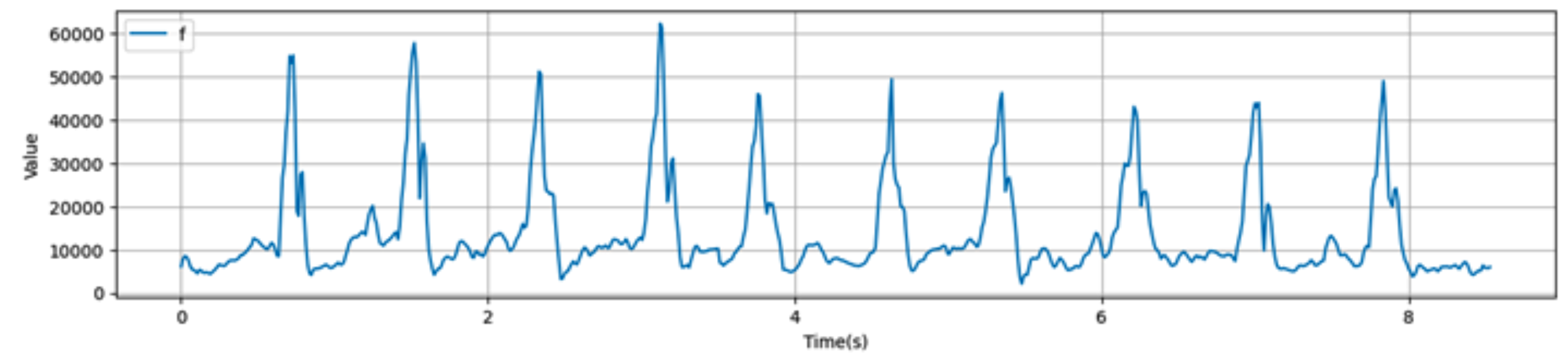}{Combined waveform from the sum of the absolute values of 3-axis accelerometer and 3-axis gyroscope.}{fig:step1}

First, we integrate the 6 waveforms from the accelerometer and gyroscope into a single $f(t)$ by summing the absolute values of these waves, as shown by Equation~\ref{eq:combine-singal}.  We call the outputted waveform the \emph{integrated waveform}.  Figure~\ref{fig:step1} displays an example of the integrated waveform.

\begin{equation}
\label{eq:combine-singal}
\begin{split}
f(t) & = \lvert A_X(t) \rvert + \lvert A_Y(t) \rvert + \lvert A_Z(t) \rvert + \\
& \lvert G_X(t) \rvert + \lvert G_Y(t) \rvert + \lvert G_Z(t) \rvert,    
\end{split}
\end{equation}
where $A_X(t)$, $A_Y(t)$, $A_Z(t)$ are the 3-axis values from accelerometer at time $t$, $G_X(t)$, $G_Y(t)$, $G_Z(t)$ are the values from gyroscope at $t$.





\fig{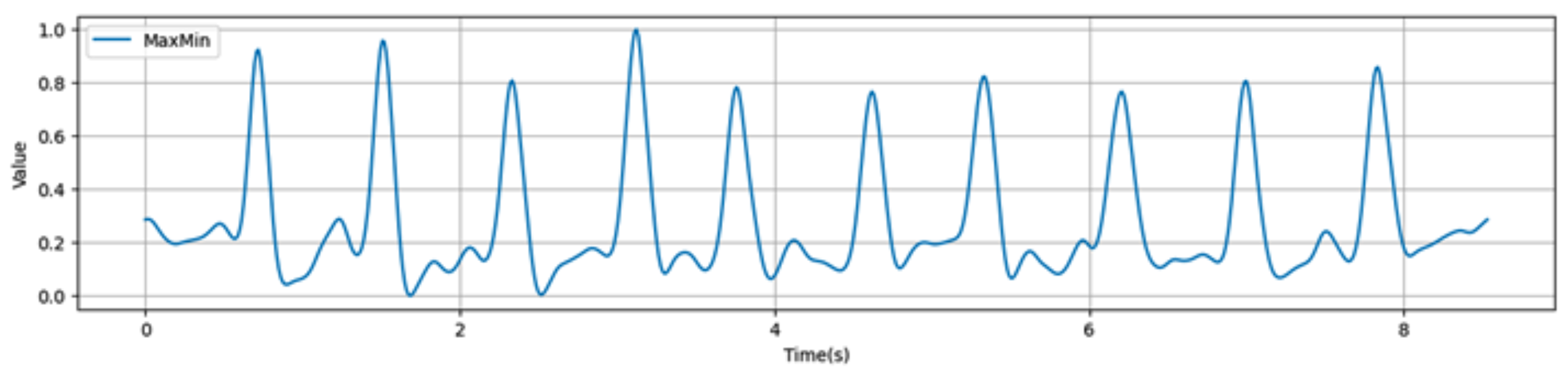}{An example of the normalized waveform.}{fig:normalized-waveform}

We normalize the integrated waveform as follows.  First, we remove the trend from the integrated waveform to remove the inconsistencies of each stroke from the same player.  Next, we apply a low-pass filter provided by ICM-20948 to remove high-frequency noise.  Finally, we scale the waveform to be within the range of 0 to 1.  These steps help to speed up the peak detection process in the subsequent steps.  Figure~\ref{fig:normalized-waveform} shows an example of the normalized waveform.

\fig{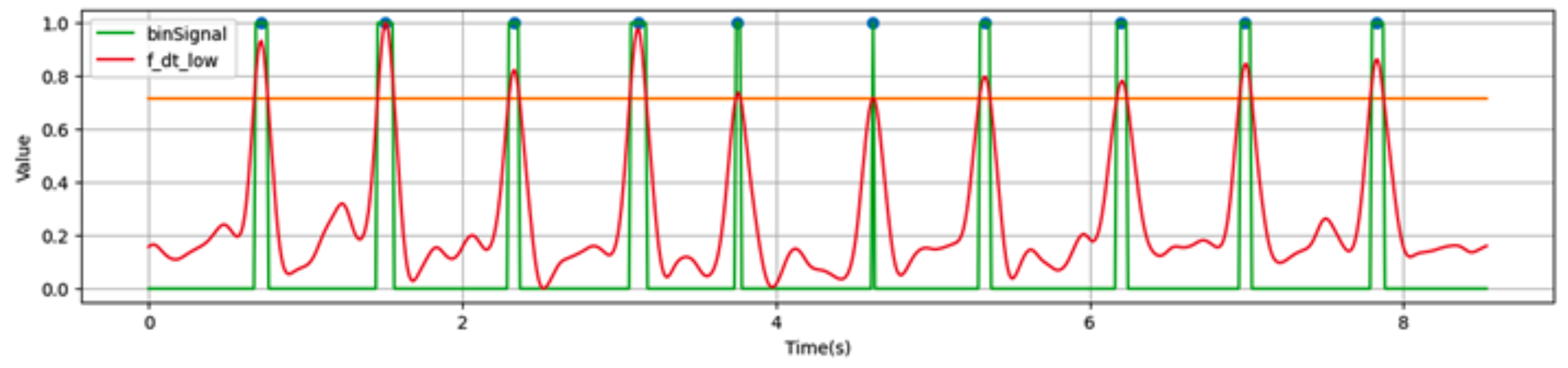}{Recursive search for the peak of each stroke waveform. In this example, given the number of peaks is 10, the search stops at $y=0.71$, in which the horizontal line intersects with the normalized waveform on 20 points.}{fig:recursive-search}

We segment the normalized waveform by stroke based on the following steps.  We first plot a horizontal line $y=1$, which interacts with the peak of the entire normalized waveform.  Since the number of swings is known, we gradually move the horizontal line down until the number of intersection points equals twice the number of swings.  For instance, in Figure~\ref{fig:recursive-search}, the number of known stroking features is 10, and the search is stopped upon finding 20 intersection points.  The peak of each stroke is expected to be within two neighboring intersections.

\fig{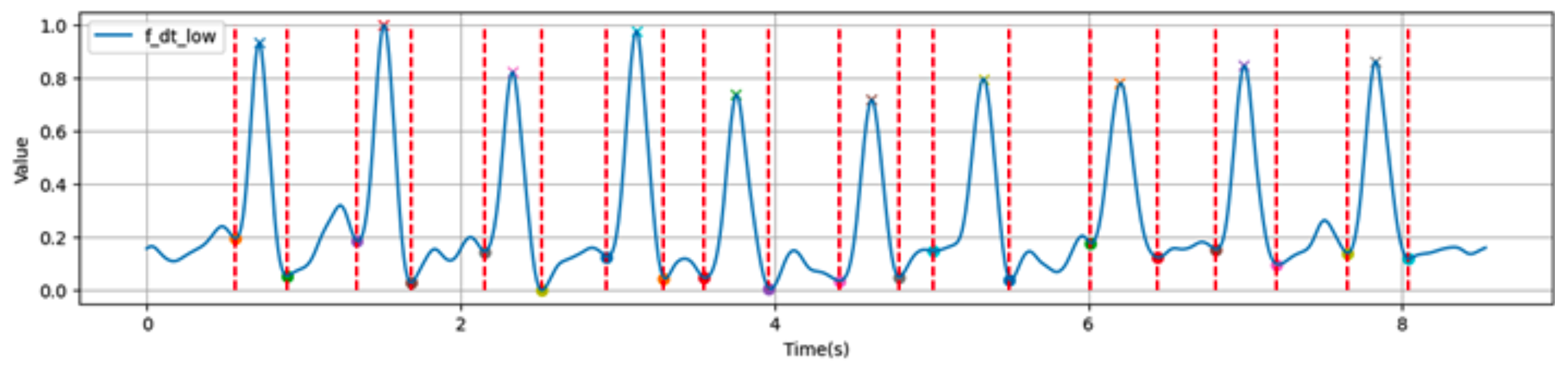}{The nearest left and right troughs of each peak is the splitting point for a single stroke waveform}{fig:troughs}

\fig{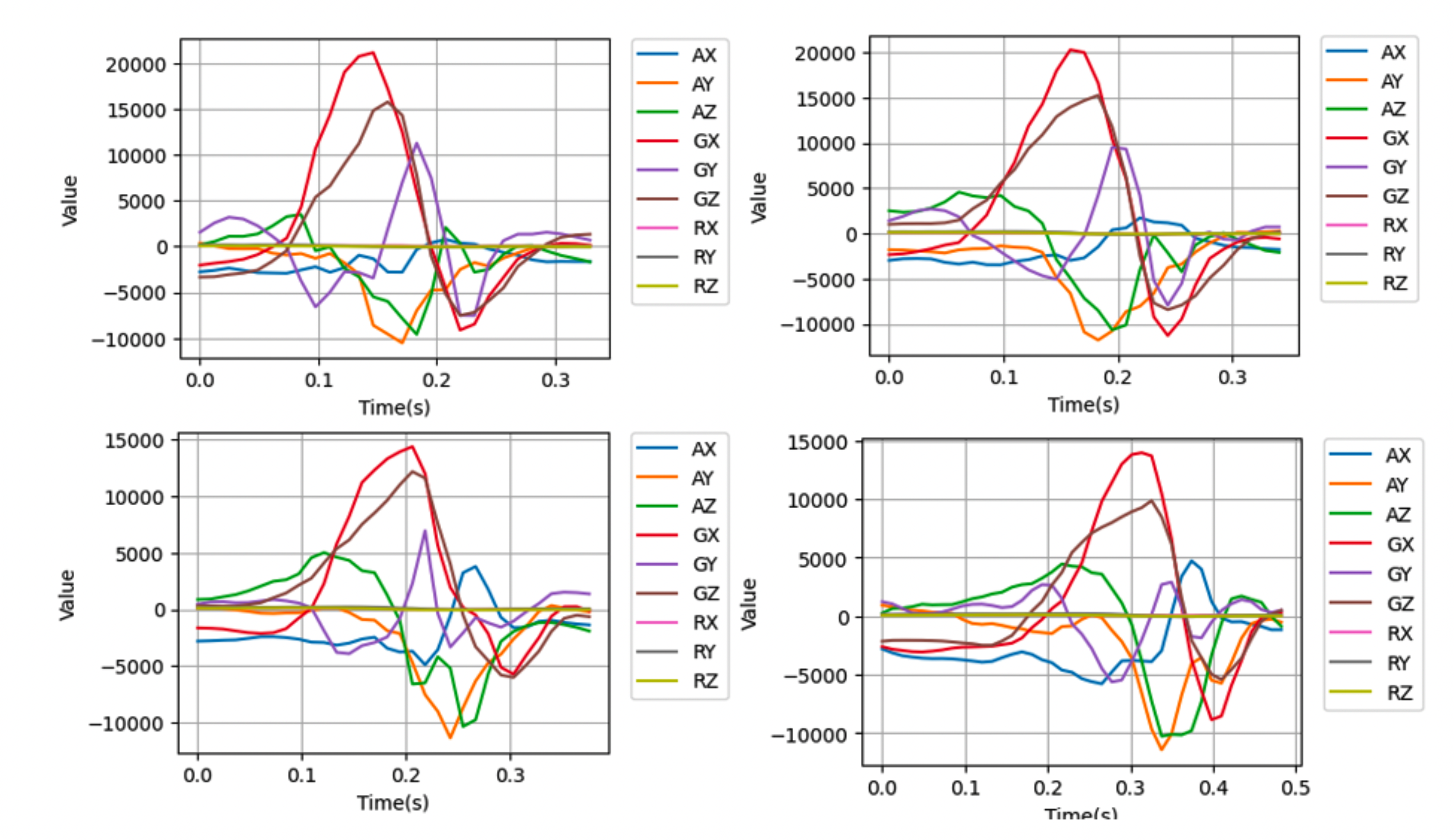}{Samples of the segmented waveforms.}{fig:final-stroke-wave}

Based on the identified peaks, we further search for the nearest troughs to the left and right. We use these two troughs as split points to separate a complete stroke waveform. Figure~\ref{fig:troughs} shows the two troughs found for each wave, and Figure~\ref{fig:final-stroke-wave} displays the segmentation results.

\subsection{Released Swing Features} \label{sec:swing-features}

\begin{table*}[tb]
\begin{tabular}{@{}lllr@{}}
\toprule
Type & \multicolumn{1}{l}{Input}                                         & \multicolumn{1}{l}{Computation}                                 & Number of generated features \\ \midrule
1    & \makecell{$A_x(t), A_y(t), A_z(t),$\\ $G_x(t), G_y(t), G_z(t)$}   & mean, variance, root mean square                              & 18 \\ \midrule
2    &  $A(t), G(t)$                                                     & mean, max, min, skewness, kurtosis                            & 10 \\ \midrule
3    &  $A(t), G(t)$                                                     & Fourier Transform, spectral density, spectral entropy & 6  \\ \bottomrule
\end{tabular}
\caption{A list of features extracted from a waveform}
\label{tab:features}
\end{table*}

Table~\ref{tab:features} displays the three types of features derived from the waveform data. The first type includes the mean, variance, and root mean square of the accelerations and angular velocities along the three axes (i.e., $A_X(t)$, $A_Y(t)$, $A_Z(t)$, $G_X(t)$, $G_Y(t)$, $G_Z(t)$), which result in 18 features. The second type contains the mean, maximum value, minimum value, skewness, and kurtosis of the overall acceleration $A(t)$ and angular velocity $G(t)$, which generate 10 features. Finally, we apply Fourier Transform to the acceleration and angular velocity waveforms and further derive the spectral density values and spectral entropy values for the acceleration and angular velocity, which result in 6 features.

By converting the continuous waveform signals into a finite number of features, it should be more convenient to apply various machine learning and deep learning models for further analyses.

\subsection{Released Personal Features}

We provide anonymized personal information for each player, including gender, age, height, weight, handedness, racket-holding hand, and years of experience. These demographic details can be utilized for group comparisons, such as examining waveform characteristics across different player groups based on factors like gender, dominant hand, or skill level.

\subsection{Data Statistics}

\begin{table}[tb]
\begin{tabular}{@{}lrrrrr@{}}
\toprule
       & \multicolumn{1}{r}{age}   & \multicolumn{1}{r}{height} & \multicolumn{1}{r}{weight} & \multicolumn{1}{r}{BMI} & exp years \\ \midrule
Min    & 12.20 & 150.0  & 35.00     & 15.42 & 1.50                \\
Q1     & 13.78 & 159.0  & 48.00     & 18.67 & 6.00                \\
Median & 15.70 & 165.0  & 56.00     & 20.24 & 7.25                \\
Mean   & 16.84 & 164.9  & 55.97     & 20.48 & 8.15                \\
Q3     & 19.70 & 170.5  & 60.00     & 22.08 & 10.00               \\
Max    & 27.40 & 183.0  & 85.00     & 31.22 & 19.00               \\ \bottomrule
\end{tabular}
\caption{Statistical summary of players' numerical features.}
\label{tab:player-num-feature-summary}
\end{table}

This section shows the statistics of the collected dataset.

We recruited 93 players, comprising 53 males and 40 females, and 78 are right-handed while 15 are left-handed.  The statistical summary for other numerical features is listed in Table~\ref{tab:player-num-feature-summary}.

Based on the swings from the 93 players, we generate more than 90,000 records after feature extraction (as discussed in Section~\ref{sec:swing-features}).  We asked each player to perform at least one of three different swing modes: swinging in the air (mode 0), full power stroke (mode 1), and stable hitting (mode 2).  The records collected for each mode are 7,500, 73,850, and 16,000, respectively.

\section{TTSWING Analysis} \label{sec:analysis}

In this section, we present pilot studies to demonstrate the value of our dataset.  Particularly, we predict users' personal information, such as gender, age, racket-holding hand, and years of experience in playing table tennis, based on the features derived from the waveforms.  We also predict a user's swing mode based on the same set of features.
It is important to note that all experiments in this section involve classification tasks, except for the regression task which focuses on predicting the number of years of experience.

\subsection{Models and Experimental Settings}

For each task, we compare the performance of 7 commonly used supervised learning algorithms: Deep Neural Network (DNN), Decision Tree (DT), k-nearest neighbors (k-NN), Support Vector Machine (SVM), Random Forest (RF), Logistic regression or Linear regression (LogReg or LR), and Na\"{i}ve Bayes (NB) or Bayesian Ridge regression (BR).

We apply the following settings for all experiments.  Prior to model training, min-max scaling is applied to ensure that all feature values are normalized between 0 and 1.  We separate the dataset into three parts by modes, except for mode classification.  Each part of the data is further split into training and test data with a ratio of 8 to 2. 

We select the important hyperparameters for different models as follows. The deep neural network includes four fully connected hidden layers with 70 neurons, 100 neurons, 100 neurons, and 10 neurons, respectively.  Each hidden layer is with the ReLU activation function.  The model is optimized using the Adam optimizer. We use the cross-entropy loss in classification and mean absolute error loss in the regression task.  We train most DNN 20 epochs with a batch size of 200 (except for the experience of years prediction because the model trained for 20 epochs is highly underfitting). The decision tree classifier utilizes the Gini impurity in classification and squared error in regression as the criterion.  For the k-Nearest Neighbor method, the value of $k$ is determined by the best performance between 1 to 20 for each task.  For Support Vector Machine, we apply the radial basis function kernel with the regularization parameter set to 1. In Random Forest, 100 base estimators are employed, with each child node containing a minimum of 10 samples. For logistic regression, we use the L2 penalty. We use the cross entropy loss for the two-class classification and softmax loss for the multi-class classification. Finally, the Gaussian Na\"{i}ve Bayes algorithm was used in Na\"{i}ve Bayes for all classification tasks, and Bayesian ridge for regression task; the shape parameter and the rate parameter for the Gamma distribution prior over the alpha parameter are
set to $1e-6$. 

We avoid extensively fine-tuning the hyperparameters to artificially enhance the results.  Instead, we mostly rely on the default settings in the employed learning libraries. However, as demonstrated later, these standard configurations yield models with satisfactory accuracies across diverse tasks. This outcome suggests that the swing data inherently encapsulates valuable information warranting further exploration.

\subsection{Gender Prediction}

\begin{table}[tb]
\centering
\begin{tabular}{@{}l|rrr@{}}
\toprule
 & \multicolumn{3}{c}{Accuracy {[}\%{]}} \\
 & \multicolumn{1}{r}{Mode0} & \multicolumn{1}{r}{Mode1} & \multicolumn{1}{r}{Mode2} \\ \midrule
DNN & 88.09 $\pm$ 1.49 & 85.18 $\pm$ 0.86 & 80.89 $\pm$ 2.18 \\
DT & 87.63 $\pm$ 0.71 & 80.71 $\pm$ 0.38 & 80.56 $\pm$ 0.66 \\
k-NN & 97.63 $\pm$ 0.31 & 91.72 $\pm$ 0.27 & 89.33 $\pm$ 0.51 \\
SVM & 88.48 $\pm$ 0.92 & 78.23 $\pm$ 0.21 & 81.84 $\pm$ 0.79 \\
RF & 94.01 $\pm$ 0.44 & 89.40 $\pm$ 0.29 & 86.78 $\pm$ 0.48 \\
LogReg & 69.07 $\pm$ 1.11 & 66.88 $\pm$ 0.33 & 70.66 $\pm$ 0.88 \\
NB & 65.20 $\pm$ 2.00 & 60.74 $\pm$ 0.32 & 54.25 $\pm$ 0.86 \\ \bottomrule
\end{tabular}
\caption{Performance of gender prediction.}
\label{tab:gender-pred-result}
\end{table}

Table~\ref{tab:gender-pred-result} shows the performance of 7 distinct classification methods across different modes. Most models perform reasonably well, with k-NN exhibiting the highest accuracy, averaging over $90\%$. This outcome indicates the presence of discernible variations between male and female swings, with waveform analysis proving capable of detecting such dissimilarities. Consequently, further investigation could delve into the specific characteristics underlying these distinctions in swing patterns.

Since the wave patterns captured by the racket sensor can effectively differentiate the gender-based differences, several interesting studies can be continued in this direction.  For example, further investigations can explore the potential influence of other factors, such as skill level or playing experience, on the detected gender-based differences in wave patterns. Additionally, perhaps exploring the underlying bio-mechanical and physiological factors and the observed waveform distinctions can provide a deeper understanding of the gender-specific characteristics in table tennis movements. 

\subsection{Age Prediction}

\begin{table}[tb]
\centering
\begin{tabular}{@{}l|rrr@{}}
\toprule
 & \multicolumn{3}{c}{Accuracy {[}\%{]}} \\
 & \multicolumn{1}{r}{Mode0} & \multicolumn{1}{r}{Mode1} & \multicolumn{1}{r}{Mode2} \\ \midrule
DNN & 78.42 $\pm$ 2.59  & 81.60  $\pm$ 1.28  & 79.26 $\pm$ 1.78        \\
DT & 86.38 $\pm$ 0.64  & 79.80  $\pm$ 0.34  & 80.83 $\pm$ 0.50         \\
k-NN & 97.39 $\pm$ 0.34  & 92.07 $\pm$ 0.28  & 90.61 $\pm$ 0.33        \\
SVM & 87.91 $\pm$ 0.78  & 77.16 $\pm$ 0.29  & 80.48 $\pm$ 0.56        \\
RF & 92.17 $\pm$ 0.64  & 87.17 $\pm$ 0.23  & 85.94 $\pm$ 0.56        \\
LogReg & 69.99 $\pm$ 0.97  & 67.86 $\pm$ 0.22  & 75.48 $\pm$ 0.57        \\
NB & 55.15 $\pm$ 1.57  & 58.41 $\pm$ 0.19  & 65.00    $\pm$ 0.02        \\ \bottomrule
\end{tabular}
\caption{Performance of age prediction.}
\label{age-pred-result}
\end{table}

In addition to gender, age is a significant factor that can potentially influence players' performance.  Indeed, age is commonly used as a classification criterion in sports competitions. Understanding the impact of age on stroke patterns could be crucial for comprehensive performance analysis. 

We classify players' ages into four groups: junior high school, senior high school, college, and adults (non-students), based on Taiwan's education system.  The classification results presented in Table~\ref{age-pred-result} show that k-NN outperforms the other methods, similar to the earlier results on gender classification. 

We may further study the relationship between age and stroke patterns in the following directions.  First, conducting a more detailed investigation within each age group may clear up the variations that may exist within specific developmental stages and may provide insights into the progression and refinement of stroke techniques over time.  Second, if we track a player's performance and stroke characteristics over several years, such longitudinal studies may reveal how the stroke patterns evolve and adapt as players gain more experience. 

\subsection{Mode Prediction}

\begin{table}[tb]
\centering
\begin{tabular}{@{}l|r@{}}
\toprule
 & Accuracy {[}\%{]} \\ \midrule
DNN                 & 97.59 $\pm$ 0.07 \\
DT       & 96.60 $\pm$ 0.15 \\
k-NN                 & 98.40 $\pm$ 0.08 \\
SVM                 & 97.32 $\pm$ 0.07 \\
RF       & 97.66 $\pm$ 0.10 \\
LogReg & 94.77 $\pm$ 0.11 \\
NB      & 92.14 $\pm$ 0.13 \\ \bottomrule
\end{tabular}
\caption{Performance of mode prediction.}
\label{tab:mode-pred-result}
\end{table}

Our dataset contains three distinct modes: swing in the air, full power stroke, and stable hitting. The results are shown in Table~\ref{tab:mode-pred-result}: all classifiers achieved an accuracy of over $92.14\%$, indicating that there is indeed some hidden information regarding the different modes within the collected waveforms.

Compared to swinging in the air, stable hitting requires a precise reaction at the moment of contact with the ball, which can potentially be reflected in the waveform characteristics. Similarly, a full power stroke involves a faster swing speed compared to stable hitting, which may result in a stronger movement and a quicker return to the ready position within the same time interval.

By utilizing a similar methodology, it is possible to differentiate various types of strokes performed by players, such as smash, loop, and push. The specific characteristics and patterns exhibited in the waveform data can provide valuable insights into the execution and effectiveness of these different stroke types. By correlating the identified stroke types with performance metrics, such as scoring efficiency or winning rates, we could better understand the impact of stroke techniques on overall player performance.

\subsection{Racket-holding hand Prediction}

\begin{table}[tb]
\centering
\begin{tabular}{@{}l|rrr@{}}
\toprule
 & \multicolumn{3}{c}{Accuracy {[}\%{]}} \\
 & \multicolumn{1}{r}{Mode0} & \multicolumn{1}{r}{Mode1} & \multicolumn{1}{r}{Mode2} \\ \midrule
DNN                 & 93.99 $\pm$ 1.06 & 94.35 $\pm$ 0.47 & 99.47 $\pm$ 0.14 \\
DT       & 93.73 $\pm$ 0.58 & 92.53 $\pm$ 0.32 & 99.92 $\pm$ 0.05 \\
k-NN                 & 98.68 $\pm$ 0.27 & 96.63 $\pm$ 0.09 & 99.73 $\pm$ 0.08 \\
SVM                 & 93.22 $\pm$ 0.60 & 90.50 $\pm$ 0.29 & 98.96 $\pm$ 0.17 \\
RF       & 95.20 $\pm$ 0.70 & 94.98 $\pm$ 0.21 & 99.89 $\pm$ 0.08 \\
LogReg & 87.36 $\pm$ 0.77 & 87.47 $\pm$ 0.24 & 92.69 $\pm$ 0.51 \\
NB      & 81.57 $\pm$ 1.01 & 83.34 $\pm$ 0.43 & 83.33 $\pm$ 1.01 \\ \bottomrule
\end{tabular}
\caption{Performance of racket-holding hand prediction.}
\label{tab:racket-holding-hand-pred-result}
\end{table}

We show the results of racket-holding hand prediction in Table~\ref{tab:racket-holding-hand-pred-result}. Despite the relatively small number of left-handed samples (15 out of 93), most classifiers achieved an accuracy rate of over $90.5\%$ in left-right hand recognition. 

Some studies have shown that left-handed players have advantages in fast-paced ball sports~\cite{loffing2017left}. In particular, in table tennis, the proportion of left-handed players in the top 100 world rankings is the second highest.  Most people believe that it is because players may have less experience in responding to left-handed shots. In addition to the experience-related reasoning, one could further explore the possibilities of other factors that may bring unique advantages to the left-handed players from waveform analyses.

\subsection{Experience in Years Prediction}

\begin{table}[tb]
\centering
\begin{tabular}{@{}l|rrr@{}}
\toprule
 & \multicolumn{3}{c}{$R^2$} \\
 & \multicolumn{1}{r}{Mode0} & \multicolumn{1}{r}{Mode1} & \multicolumn{1}{r}{Mode2} \\ \midrule
DNN & 0.49 $\pm$ 0.44  & 0.86 $\pm$ 0.04  & 0.80 $\pm$ 0.01 \\
DT & 0.90 $\pm$ 0.02  & 0.86 $\pm$ 0.01  & 0.87 $\pm$ 0.01 \\
k-NN & 0.97 $\pm$ 0.01  & 0.94 $\pm$ 0.00  & 0.92 $\pm$ 0.01 \\
SVM & 0.70 $\pm$ 0.02  & 0.63 $\pm$ 0.01  & 0.78 $\pm$ 0.01 \\
RF & 0.94 $\pm$ 0.01  & 0.93 $\pm$ 0.00  & 0.93 $\pm$ 0.00 \\
LR & 0.47 $\pm$ 0.03  & 0.49 $\pm$ 0.01  & 0.73 $\pm$ 0.01 \\
BR & 0.47 $\pm$ 0.03  & 0.49 $\pm$ 0.01 & 0.73 $\pm$ 0.01 \\ \bottomrule
\end{tabular}
\caption{Performance of experience in years prediction.} 
\label{tab:exp-years-pred-result}
\end{table}

In addition to the 34 waveform features, we also incorporate the age feature in this experiment due to the significance of age in predicting the years of experience in playing table tennis. Specifically, it is reasonable to assume that a player who is $y$ years old cannot have more than $y$ years of experience. Therefore, considering age as a feature helps us predict experience in years.

Table~\ref{tab:exp-years-pred-result} provides the $R^2$ score in predicting the years of experience. Except for Linear Regression and Bayesian Ridge regression, all other methods perform reasonably well. The result suggests that players with varying years of experience may exhibit distinct swing behaviors. These findings open up opportunities for further exploration of the relationship between experience, swing behaviors, and performance outcomes in table tennis.
\section{Conclusion and Future Work} \label{sec:disc}

This paper presents an open dataset for table tennis analyses.  The dataset is unique in several ways.  First, we collect the dataset from elite players.  Second, we include the swing and players' anonymized demographic information.  Finally, to the best of our knowledge, this is the largest dataset in terms of both the number of involved players and the number of collected strokes.  We will continue collecting more players' swing information and perhaps other table tennis-related features to enrich the dataset. Besides, we conducted pilot studies on this dataset to show its effectiveness.  The results demonstrate the successful prediction of several important attributes of table tennis players, including gender, age, swing force, racket-holding hand, and years of experience. These findings open up promising avenues for future research in table tennis analysis and player profiling.

One potential research direction is to explore the relationship between the predicted attributes and player performance in table tennis. By analyzing the gender-based performance differences, the influence of age on playing style and effectiveness, and the impact of swing force and racket-holding hand on player strategies and outcomes, we may obtain insights into the factors contributing to success in table tennis. 

Another area for future investigation is optimizing training programs and skill development strategies based on players' attributes. By tailoring training methods to specific player characteristics, such as age and experience, we can improve the effectiveness of training programs and accelerate skill acquisition in table tennis. 

Furthermore, the predicted attributes can be used for player classification and talent identification. Developing classification models that automatically categorize players into different skill levels or playing styles can aid in talent identification. Coaches can utilize these models to identify promising players based on their predicted attributes to uncover hidden talents.

Injury prevention and rehabilitation are also important areas to explore. For example, investigating the relationship between player characteristics, such as age and swing force, and the risk of injuries may help reduce the occurrence and severity of injuries in table tennis.

Finally, we would also like to analyze different swing movements, such as forehand topspin and backhand stroke, to understand these intricate skills in greater depth. Such a study may help researchers and coaches unravel the complexities contributing to a player's overall performance.  This analysis not only enhances our understanding of different swing movements but also helps the development of training methodologies, coaching strategies, and equipment designs to nurture the next generation of elite table tennis players.

In conclusion, our dataset and studies open up exciting possibilities for future research in table tennis analysis. We can advance the table tennis field by further investigating the relationship between player attributes and various aspects of the sport, such as performance, training, classification, injury prevention, and player experience, ultimately benefiting players, coaches, and the sport.

%
%
%
\section*{Acknowledgments}
We acknowledge support from the National Science and Technology Council of Taiwan under grant number NSTC 112-2425-H-028-001.

%

\bibliographystyle{named}
\bibliography{ref}

\end{document}